\documentclass[lettersize,journal]{IEEEtran}
\usepackage{amsmath,amsfonts}
\usepackage{algorithmic}
\usepackage{algorithm}
\usepackage{array}
\usepackage[caption=false,font=normalsize,labelfont=sf,textfont=sf]{subfig}
\usepackage{textcomp}
\usepackage{stfloats}
\usepackage{url}
\usepackage{verbatim}
\usepackage{graphicx}
\usepackage{hyperref}
\usepackage{tikz}
\usepackage[normalem]{ulem}
\usepackage{biblatex}
\bibliography{references}
\usetikzlibrary{shapes, arrows.meta, quantikz2, positioning}
\usepackage{booktabs} 

\begin{document}

\title{Quantum Gated Recurrent GAN with Gaussian Uncertainty for Network Anomaly Detection}

\author{Wajdi Hammami,~\IEEEmembership{Member,~IEEE},
        Soumaya Cherkaoui,~\IEEEmembership{Senior Member,~IEEE}, 
        Jean-Frederic Laprade,
        Ola Ahmad,
        Shengrui Wang

\thanks{ Wajdi Hammami and Soumaya Cherkaoui are with the Department
of  Computer and Software Engineering,  Polytechnique Montréal, Montréal, Canada, e-mail: (wajdi.hammami@polymtl.ca).}
\thanks{Jean-Frederic Laprade is with the Institut Quantique,  Sherbrooke, Canada.} 
\thanks{Ola Ahmad is with Thales Digital Solutions Canada}
\thanks{Shengrui Wang is with the Université de Sherbrooke, Sherbrooke, Canada.}

}

\markboth{Journal of \LaTeX\ Class Files,~Vol.~14, No.~8, August~2021}%
{Shell \MakeLowercase{\textit{et al.}}: A Sample Article Using IEEEtran.cls for IEEE Journals}


\maketitle

\begin{abstract}
Anomaly detection in time-series data is a critical challenge with significant implications for network security. Recent quantum machine learning approaches, such as quantum kernel methods and variational quantum circuits, have shown promise in capturing complex data distributions for anomaly detection but remain constrained by limited qubit counts. We introduce in this work a novel Quantum Gated Recurrent Unit (QGRU)-based Generative Adversarial Network (GAN) employing Successive Data Injection (SuDaI) and a multi-metric gating strategy for robust network anomaly detection. 
Our model uniquely utilizes a quantum-enhanced generator that outputs parameters (mean and log-variance) of a Gaussian distribution via reparameterization, combined with a Wasserstein critic to stabilize adversarial training. Anomalies are identified through a novel gating mechanism that initially flags potential anomalies based on Gaussian uncertainty estimates and subsequently verifies them using a composite of  critic scores and reconstruction errors. Evaluated on benchmark datasets, our method achieves a high time-series aware F1 score (TaF1) of 89.43\% demonstrating superior capability in detecting anomalies accurately and promptly as compared to existing classical and quantum models. Furthermore, the trained QGRU-WGAN was deployed on real IBM Quantum hardware, where it retained high anomaly detection performance, confirming its robustness and practical feasibility on current noisy intermediate-scale quantum (NISQ) devices.
\end{abstract}

\begin{IEEEkeywords}
Network anomaly detection, Generative adversarial networks, Hybrid quantum-classical models, Industrial control systems, Quantum GRU, Quantum machine learning, Successive data injection, Time-series forecasting, Variational quantum circuits, Wasserstein GAN.
\end{IEEEkeywords}

\section{Introduction}
\label{sec:introduction}
\IEEEPARstart{A}{nomaly} detection in time-series data plays a vital role in monitoring the behavior of complex and dynamic communication systems, where temporal dependencies heavily influence traffic patterns ~\cite{ahmed2016survey}.
These anomalies—subtle and often transient deviations from normal network behavior—can indicate serious issues such as cyberattacks or system failures. Unlike static datasets, time-series network data presents unique challenges, since irregularities must not only be flagged, but also explained within the evolving temporal context of network activity.
Traditional statistical approaches often fall short in capturing non-linear dependencies or adapting to evolving patterns over time, necessitating more sophisticated learning-based methods to model the temporal and structural intricacies of modern networks.

In recent years, machine learning has become a cornerstone in solving network anomaly detection problems across various domains \cite{zhang2019network}. By learning patterns directly from data, machine learning models can adapt to complex, high-dimensional environments where manual rule-setting is impractical. Techniques such as autoencoders \cite{an2015variational}, recurrent neural networks \cite{malhotra2015lstm}, and generative models \cite{hammami2025quantum} have shown strong capabilities in modeling normal behavior and flagging deviations. These models are particularly valuable in settings where data distributions shift over time or labeled anomalies are scarce. As a result, machine learning based 
anomaly detection techniques has become integral component for modern network security systems~\cite{landauer2025review}.

Alongside the advancements in classical machine learning, quantum computing has emerged as a promising frontier for solving complex computational problems. Quantum machine learning (QML), a fusion of quantum computing principles with data-driven learning, is attracting increasing attention for its potential to outperform classical models in specific tasks \cite{hdaib2024quantum}, thanks to its leveraging quantum phenomena such as superposition and entanglement to encode rich feature spaces and capture complex correlations that are challenging for classical models.
While empirical evidence on large-scale real-world datasets remains limited, recent studies suggest that hybrid quantum-classical models may offer advantages in scenarios involving high-dimensional data, limited samples, or probabilistic modeling \cite{dunjko2018machine} ---challenges commonly encountered in network anomaly detection tasks. This growing synergy between quantum computing and machine learning opens new avenues for addressing the complexity of network anomaly detection in ways that classical methods may struggle to achieve. 


In this work, we focus on the problem of anomaly detection within network time-series data—a setting where temporal structure plays a crucial role in identifying irregularities. Given the sequential nature of the data as well as the need to capture contextual patterns and handle uncertainty, we propose to leverage the strengths of both classical machine learning and quantum computing. Specifically, we explore the integration of recurrent neural architectures, known for their temporal modeling capabilities, with quantum-enhanced learning mechanisms that offer expressive generative potential~\cite{jerbi2023quantum}. This convergence allows us to build a hybrid model tailored for detecting anomalies in network time-series, where subtle and context-dependent deviations must be identified with high precision.

We address several key limitations that arise in time-series network anomaly detection. While traditional GAN architectures can model data distributions, they often struggle to capture complex temporal dependencies inherent in sequential data. To address this limitation, we integrate Gated Recurrent Units (GRUs), which are well-suited for modeling sequential patterns due to their efficient parameterization. To enrich the model’s expressive capacity, we introduce Variational Quantum Circuits (VQCs) that embed the latent representations into richer Hilbert spaces, enabling the capture of complex feature interactions \cite{benedetti2019parameterized}. Since directly encoding large input dimensions into quantum circuits remains constrained by qubit availability, we employ Successive Data Injection \cite{kalfon2024sudai} (SuDaI), which allows multiple segments of temporal data to be sequentially injected into the quantum circuit, thereby efficiently utilizing limited quantum resources. Finally, we combine these components within a hybrid GAN architecture, and introduce a multi-metric anomaly detection pipeline that leverages predictive uncertainty, adversarial feedback, and reconstruction-based criteria to robustly detect anomalies with greater contextual sensitivity. 


We have evaluated our method using the HAI (HIL-based Augmented Industrial Control Systems) dataset \cite{shin2020hai}, which provides multivariate time-series data collected from a network-integrated industrial control system under both normal operations and staged network-based cyber-attacks. Performance is assessed using the extended time-series aware F1 score (TaF1) metric \cite{hwang2022accuracy}, which accounts for anomaly detection accuracy over temporal intervals. Our approach achieves an TaF1 score of 88.63\%, substantially surpassing existing methods. Furthermore, the trained model was successfully executed on real IBM Quantum hardware without retraining, demonstrating that the proposed quantum architecture is not only effective in simulation but also operationally viable on current NISQ devices.


The paper proceeds as follows: Section~\ref{sec:related-work} covers related literature in classical and quantum anomaly detection methods. Section~\ref{sec:proposed-method} outlines our proposed model's architecture. Section~\ref{sec:results} details the experimental design and results. Section~\ref{sec:conclusion} concludes the paper.

\section{Related works} \label{sec:related-work}
Time-series anomaly detection (TSAD) has gained significant attention across domains such as cybersecurity, industrial monitoring, and IoT systems \cite{chandola2009anomaly}, where early detection of irregular behavior is crucial. Traditional statistical and clustering methods \cite{chatfield2003analysis} offer basic capabilities but struggle with high-dimensional, non-linear, and noisy data. Deep learning has advanced the field through generative and prediction-based models, particularly by using RNNs and GANs \cite{lin2020idsgan, dimattia2019gans, wang2025anomaly, zhang2019network}. These models have shown success in capturing complex temporal patterns but often face challenges related to data imbalance, threshold setting, and explainability \cite{saravanan2023tsigan, zhang2019unsupervised}.

GAN-based models~\cite{zhao2024gan_nids} represent a prominent approach in unsupervised anomaly detection for network systems. These models learn the distribution of normal data and detect anomaly by measuring deviations. Despite recent progresses, stabilizing GAN training remains challenging due to issues such as mode collapse and unstable convergence. Methods such as the Wasserstein loss  \cite{haloui2018wgan} have been proposed to address these challenges, allowing to improve the training stability without fully resolving the problem. Simultaneously, data augmentation through generative models aids in addressing the rarity of anomalies by synthesizing diverse samples \cite{lim2018doping}.


Quantum machine learning (QML) introduces new tools for anomaly detection while leveraging the computational advantages of quantum systems to model complex data more efficiently. Quantum Recurrent Neural Networks, particularly Quantum GRUs (QGRUs), have shown improved generalization and training efficiency~\cite{defalco2024hybrid}. Quantum GANs (QGANs) combine quantum circuits with generative models, enabling efficient sampling and representation of high-dimensional distributions \cite{hammami2024quantumgan}. Strategies such as SuDaI and data re-uploading address hardware constraints \cite{kalfon2024sudai, perez2019reuploading}, while hybrid quantum-classical frameworks enable deployment on current Noisy Intermediate-Scale Quantum (NISQ) devices \cite{alhussein2023hybrid}.

Finally, evaluating TSAD models requires appropriate metrics that account for temporal structures. Traditional point-based F1 scores often misrepresent performance and may significantly overrate detection effectiveness, especially in time-dependent settings \cite{hwang2022overrate}. As a response to these limitations, time-aware metrics such as TaF1 \cite{tatbul2018precision} and PATE \cite{ghorbani2024pate} have emerged to evaluate detection accuracy over intervals, allowing to account for early or delayed responses and to better reflect real-world utility. Our adoption of TaF1 aligns with these modern evaluation standards, ensuring that our model’s performance is meaningfully measured.

\section{Proposed Method}\label{sec:proposed-method}
\subsection{Problem Formulation}

Let \( X = [\boldsymbol{x}_1, \boldsymbol{x}_2, \dots, \boldsymbol{x}_T] \) with \( \boldsymbol{x}_t \in \mathbb{R}^{d} \) denote a multivariate time-series consisting of \( T \) time steps and \( d \) features. 
The goal is to detect whether a given point \( x_t \) at time \( t \) is anomalous based on the context provided by a preceding window of size \( w \). In our unsupervised setting, the training data is assumed to contain predominantly normal patterns, and no anomaly label is used during model training.

We define the input to the model to be a sliding window \( W_t = [x_{t-w}, \dots, x_{t-1}] \), which conditions the model to predict the next point \( \hat{x}_t \). Deviations between the predicted distribution and the actual observation are often used to compute anomaly scores. Since real-world time series often exhibit temporal drift, subtle deviations, or bursty behaviors, anomalies may arise under changing conditions and be difficult to detect consistently. As a result, the model must capture not only the expected value, but also the associated uncertainty and the contextual consistency of each prediction.
Therefore, we formulate anomaly detection as a conditional sequence modeling task: the model learns a conditional distribution \( p(x_t | W_t) \), and decides whether \( x_t \) is an anomaly depending on whether it significantly deviates from this distribution according to a composite scoring function.

\subsection{Data Preparation}

Our experiments are conducted on version 21.03 of the HAI dataset~\cite{shin2020hai}, which consists of multivariate time-series data collected from a network-integrated industrial control system (ICS) operating under both normal and attack conditions. Each sample corresponds to sensor and actuator readings comprising 86 features that reflect the physical and network states of the system. The dataset spans approximately 25.5 days of recorded time-series data, including around 20 days of normal operation and 5.5 days under attack scenarios. Data are recorded at a frequency of 59 measurements per second. The anomaly events in the dataset correspond to 38 cyberattacks generated using 14 distinct attack primitives --elementary malicious actions such as Set Point (SP) manipulation, Control Output (CO) override, or Process Variable (PV) spoofing--. These attacks vary in duration from 154 to 2880 seconds and range from simple one-primitive intrusions to more complex, stealthy, or combined attacks. Notably, the dataset includes 15 stealth attacks designed to conceal abnormal behaviors by replaying pre-recorded PV values, as well as gradual attacks exhibiting linear transitions (temporal drift) and discrete bursts of activity. This diversity makes the HAI dataset particularly challenging and realistic for network anomaly detection in ICS environments.



The dataset is divided into a training set and a test set. One-third of the test set is used as a validation set for feature selection and hyperparameter tuning, while the remaining two-thirds are held out and kept completely unseen for final performance evaluation.

\begin{figure}[h]
\centering
\includegraphics[width=0.9\linewidth]{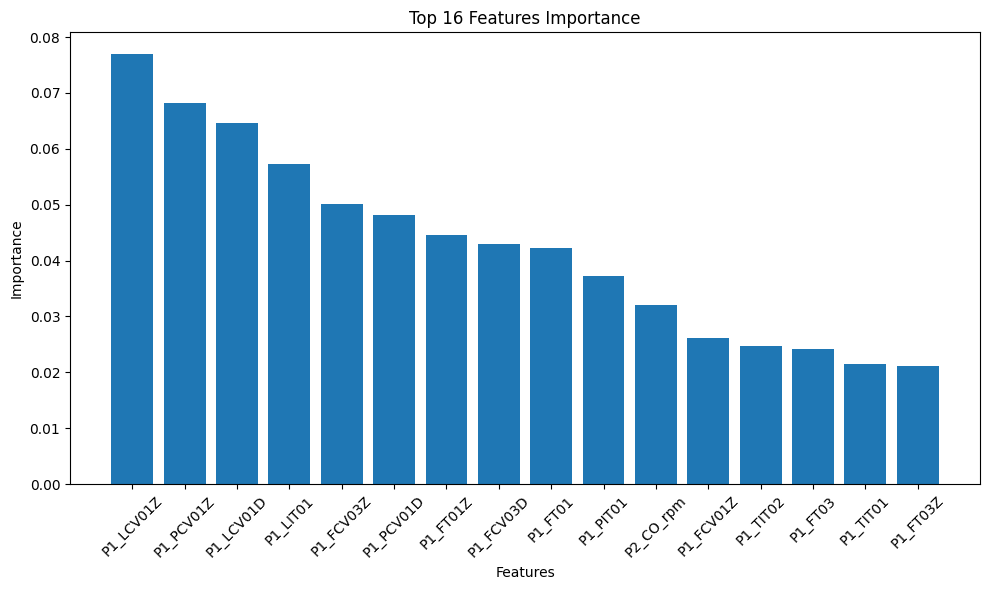}
\caption{Top 16 Feature Importances ranked by Gini criterion. These features were selected for their strong relevance to anomaly detection and used as model inputs to reduce dimensionality.}
\label{fig:feature_importance}
\end{figure}

To reduce dimensionality and improve learning efficiency, we applied feature selection using the Gini importance criterion, following the same methodology as proposed in \cite{cultice2024anomaly}, which was conducted on the same dataset.
This method evaluates the relevance of each feature by measuring its ability to reduce uncertainty across classification splits. The top 16 ranked features are retained, capturing the most informative aspects.  
Figure~\ref{fig:feature_importance} illustrates the results of our feature selection.
We begin by preprocessing the training set using MinMax normalization to scale all features to the range [0, 1]. Subsequently, the normalized data is segmented into fixed-length windows of size \(w\) with no overlap, enabling the model to capture local temporal patterns.

For a time-series of length \(T\), our dataset is therefore composed of \(\left \lfloor{T / w}\right \rfloor \) inputs, a number that can be excessive given the high computational cost of simulating and training quantum circuits. We employ a clustering-based downsampling strategy to reduce the size of the training set while preserving the diversity of temporal patterns. Each time-series window is flattened into a feature vector and the ensemble of the feature vectors is clustered using K-Means. Each window is therefore assigned to a cluster and the centroid of each cluster is used as a representative sample. Then, the centroid of each cluster is used to represent all windows assigned to that cluster. This results in a more compact yet representative training set, well-suited for efficient quantum model training. The number of clusters is chosen based on validation performance to strike a balance between reduction and pattern preservation.


\subsection{Model Overview}

In this section, we first introduce the Hybrid Quantum Layers (HQLs), the fundamental building blocks of our model. We then describe the Quantum-Enhanced GRU (QGRU) architecture, followed by the overall design of our generative adversarial network (GAN).

\subsubsection{Hybrid Quantum Layers (HQLs)}

HQLs have been designed to serve as the core quantum computational units in our model, integrating a classical preprocessing stage, a quantum variational circuit, and a classical postprocessing stage. Each HQL consists of a classical linear transformation, a Variational Quantum Circuit (VQC), and a final classical projection layer. The initial dense layer projects the GRU input into a lower-dimensional latent space, which is then injected into the VQC using a  strategy known as \textit{Successive Data Injection} (SuDaI) that we have developed recently~\cite{kalfon2024sudai}. In this scheme, the input vector is not injected all at once but introduced progressively across the depth of the circuit. Specifically, the vector components are encoded as rotation angles of single-qubit gates (typically along the \( R_y \) or \( R_z \) axes) and applied sequentially at multiple circuit layers. In our model, we adopt a SuDaI-inspired scheme not to encode temporal information, but to progressively inject different components of a high-dimensional classical representation (e.g., the output of a dense layer) into a limited number of qubits. This approach enables us to incrementally construct a rich quantum state while staying within hardware constraints.

\begin{figure*}[t]
\centering
\resizebox{\textwidth}{!}{%
\begin{tikzpicture}[node distance=0.8cm and 0.5cm, every node/.style={font=\small}, >=Stealth]

\node[draw, circle, minimum size=0.6cm] (in1) at (0,-1) {$x_{1}$};
\node[draw, circle, minimum size=0.6cm] (in2) at (0,-4) {$x_{4}$};
\node[above=0.3cm of in1] {Classical Input Layer};

\foreach \t in {0.4,0.5,0.6} {
  \node[fill, circle, inner sep=1pt, black]
    at ($(in1)!\t!(in2)$) {};
}

\node[ minimum size=0.6cm, right=1cm of in1, yshift={(1-1.6)*1cm}] (c1) {$h_{1}$};
\node[ minimum size=0.6cm, right=1cm of in1, yshift={(1-3.4)*1cm}] (c2) {$h_{6}$};
\foreach \t in {0.4,0.5,0.6} {
  \node[fill, circle, inner sep=1pt, black]
    at ($(c1)!\t!(c2)$) {};
}

\foreach \j in {1,...,2} {
  \foreach \i in {1,...,2} {
    \draw[->, gray] (in\i) -- (c\j);
  }
}

\node (vqc) [right=1.4cm of in1,  yshift={(1-2.3)*1cm}] {
  \begin{quantikz}[row sep={0.6cm,between origins}, column sep=0.3cm]
    \lstick{$q_1$} & \gate{Ry(arctan(h_1))} & \gate{Rz(\theta_1^1)} & \gate{Rx(\theta_2^1)} & \ctrl{1}          & \qw          & \gate{Ry(arctan(h_4))} & \gate{Rz(\theta_1^2)} & \gate{Rx(\theta_2^2)} & \ctrl{1}          & \qw          &  \meter{}          \\
    \lstick{$q_2$} & \gate{Ry(arctan(h_2))} & \gate{Rz(\theta_3^1)} & \gate{Rx(\theta_4^1)} & \targ{}          & \ctrl{1}     & \gate{Ry(arctan(h_5))} & \gate{Rz(\theta_3^2)} & \gate{Rx(\theta_4^2)} & \targ{}          & \ctrl{1}       & \meter{}          \\
    \lstick{$q_3$} & \gate{Ry(arctan(h_3))} & \gate{Rz(\theta_5^1)} & \gate{Rx(\theta_6^1)} & \qw              & \targ{}      & \gate{Ry(arctan(h_6))} & \gate{Rz(\theta_5^2)} & \gate{Rx(\theta_6^2)} & \qw              & \targ{}       & \meter{}          
  \end{quantikz}
};

\foreach \i in {1,...,3}{
\foreach \k in {1,2,3,4} {
  \node[draw, circle, minimum size=0.5cm, right=3cm of vqc] (m\k) at (16, -\k ) {$y_\k$};
  \draw[->, gray] (vqc.east|-vqc.south)+(0 , 2.1cm-\i*0.5cm) -- (m\k.west);
}
}
\node[above=0.3cm of m1] {Classical Postprocessing Layer};
\end{tikzpicture}
}
\caption{Architecture of the Hybrid Quantum Layer (HQL). A small set of classical input neurons is first mapped through a fully connected classical layer. The resulting activations are then successively injected into the rotation angles of single qubit gates. Each qubit undergoes a sequence of parameterized rotations (e.g., $R_y$, $R_z$, $R_x$), followed by an entangling layer composed of CNOT gates. This structure is repeated across multiple layers of the variational quantum circuit (VQC). The quantum state is measured at the end of the circuit, and the expectation value of the Pauli-$Z$ observable is computed for each qubit. These values are then passed through a lightweight classical post-processing layer to produce the final output.  Gray arrows indicate the direction of data flow.}

\label{fig:hql-diagram}
\end{figure*}
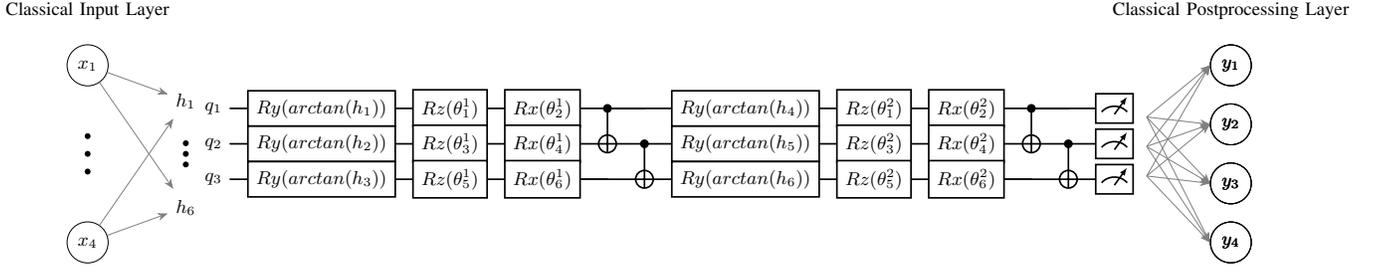

The quantum circuit’s output is measured by computing the expectation value of the Pauli-Z observable on each qubit individually, resulting in one scalar output per qubit. These expectation values are used as the quantum layer’s output and are optimized jointly with the rest of the model. 
and passed through a final classical linear layer to produce the output. This hybrid design enables the model to exploit both classical and quantum representations, thereby enhancing its ability to capture complex, non-linear temporal dependencies while remaining compatible with NISQ devices. Figure~\ref{fig:hql-diagram} illustrates the architecture of the HQL.

\subsubsection{Quantum-Enhanced GRU (QGRU)}
The  QGRU is a recurrent unit in which the traditional gating mechanisms—update, reset, and candidate gates—are implemented by HQLs. 

At each time step, the QGRU takes the current input vector and the previous hidden state and processes them through HQLs to compute the update and reset gate values. The update gate determines how much of the past hidden state should be retained, while the reset gate controls the influence of the previous state in generating the new candidate activation \cite{chung2014gru}. The quantum component consists of Variational Quantum Circuits (VQCs), where classical input is encoded via angle encoding and each qubit's output is computed as the expectation value of the Pauli-\(Z\) observable. 

Our architecture draws inspiration from the hybrid QGRU proposed by De Falco et al.~\cite{defalco2024hybrid}, which also interfaces classical dense layers with quantum circuits to implement gating mechanisms. However, unlike their design that includes both an input and an output fully connected layer surrounding each VQC (FC\textsubscript{in} \(\rightarrow\) VQC \(\rightarrow\) FC\textsubscript{out}), we employ only a single classical-to-quantum interface layer before the VQC. This simplification reduces the number of trainable parameters while maintaining the expressive power of the quantum gates. Furthermore, instead of encoding the entire input at once, we inject different components of a high-dimensional vector across successive quantum layers, allowing us to build a richer quantum state using a limited number of qubits.




\subsubsection{Quantum Wasserstein Generative Adversarial Network (QWGAN)}

Generative Adversarial Networks (GANs) are composed of two competing neural networks: a generator and a discriminator (or critic). The generator learns to produce data that mimics the real distribution, while the critic attempts to distinguish between real and generated samples. Through this adversarial training, the generator improves by minimizing the discrepancy between generated and real samples. In our setting, the generator aims to model the conditional distribution of the next time-step given a sequence of previous observations, making it suitable for time-series forecasting and anomaly detection tasks. The adversarial nature of GANs enables learning complex distributions in an unsupervised manner, a key advantage in settings where labeled anomalies are scarce or unavailable which is the case here as the dataset training set contain only bengin data.

\begin{figure*}[h]
\centering
\begin{tikzpicture}[
    node distance=0.1cm and 0.3cm,
    every node/.style={font=\small},
    arrow/.style={-Stealth, thick},
    block/.style={rectangle, draw, rounded corners, minimum width=2cm, minimum height=0.8cm, fill=blue!20},
    io/.style={circle, draw, minimum size=0.8cm, fill=orange!20}
  ]

  \node[io] (win) {{$X_{t-w+1:t}$}};
  \node[block, right=of win] (gruG) {QGRU};
  \node[block, above right=of gruG] (hqlG1) {HQL};
  \node[block, below right=of gruG] (hqlG2) {HQL};
  \node[io, right=of hqlG1] (mu) {{$\mu_t$}};
  \node[io, right=of hqlG2] (logvar) {{$\log\sigma^2_t$}};
  \node[block, right=1cm of $(mu)!0.5!(logvar)$] (reparam) {Reparam};
  \node[io, right=of reparam] (xfake) {{$\hat x_{t}$}};
  \node[below=1cm of $(gruG)!0.5!(hqlG2)$, font=\bfseries] (Generator) {Generator};

  \draw[arrow] (win) -- (gruG);
  \draw[arrow] (gruG) -- (hqlG1);
  \draw[arrow] (gruG) -- (hqlG2);
  \draw[arrow] (hqlG1) -- (mu);
  \draw[arrow] (hqlG2) -- (logvar);
  \draw[arrow] (mu) -- (reparam);
  \draw[arrow] (logvar) -- (reparam);
  \draw[arrow] (reparam) -- (xfake);

  \node[io, above=of xfake] (xreal) {{$x_t$}};
  \node[block, right=1cm of xreal] (gruC1) {QGRU};
  \node[block, right=of gruC1] (hqlC1) {HQL};
  \node[io, right=of hqlC1] (score1) {score};

  \node[block, right=1cm of xfake] (gruC2) {QGRU};
  \node[block, right=of gruC2] (hqlC2) {HQL};
  \node[io, right=of hqlC2] (score2) {score};
  
  \node[below=0.8cm of $(hqlC2)!0.5!(gruC2)$, font=\bfseries] (crit) {Critic input: $X_{t-w+1:t} \parallel \{x_t, \hat{x}_t\}$};
  \node[below=1.4cm of $(hqlC2)!0.5!(gruC2)$, font=\bfseries] (crit) {Critic};

  \draw[arrow] (xreal) -- (gruC1);
  \draw[arrow] (xfake) -- (gruC2);
  \draw[arrow] (gruC1) -- (hqlC1);
  \draw[arrow] (gruC2) -- (hqlC2);
  \draw[arrow] (hqlC1) -- (score1);
  \draw[arrow] (hqlC2) -- (score2);

\end{tikzpicture}
\caption{Model overview: the generator (left) uses a QGRU backbone followed by two Hybrid Quantum Layers (HQLs) to produce Gaussian parameters $\mu_t$ and $\log\sigma^2_t$, applies the reparameterization trick, and yields a sample $\hat x_{t}$. The critic (right) processes both the real next point $x_{t}$ and generated sample through parallel QGRU+HQL pipelines to compute Wasserstein scores. Colored blocks distinguish classical (orange) and quantum-augmented (blue) components.}
\label{fig:model-overview} 
\end{figure*}
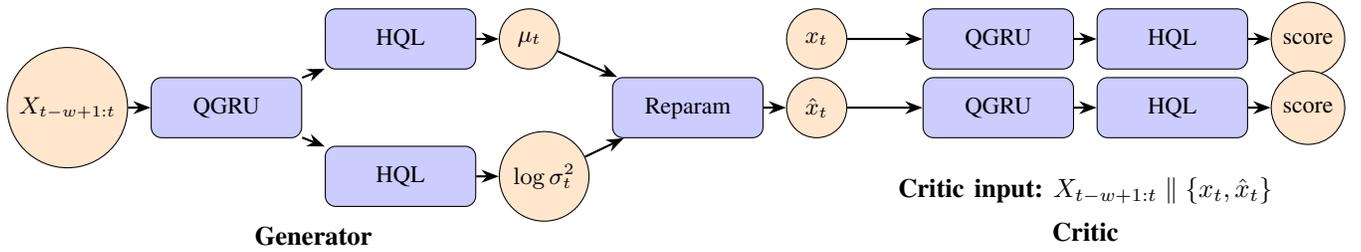

The generator works in such a way that at each time step, it receives a fixed-length sliding window of historical multivariate data \( W_t = [x_{t-w}, \ldots, x_{t-1}] \) and is tasked with forecasting the next time step. This input is first processed by a QGRU to extract sequential features, which are then passed through two parallel HQLs. Each HQL produces one of the parameters of a Gaussian distribution: the mean \( \mu_t \) and the log-variance \( \log(\sigma_t^2) \). This design enables the generator to learn a full predictive distribution over the next time step, capturing both the expected value and the associated uncertainty.

Following that, the model generates a sample from the predicted Gaussian distribution using the reparameterization trick~\cite{kingma2014auto}:
\[
\hat{x}_t = \mu_t + \sigma_t \cdot \epsilon, \quad \epsilon \sim \mathcal{N}(0, I)
\]
This formulation expresses the sampling operation as a differentiable transformation of a noise term, which enables gradient-based optimization despite the presence of stochasticity. In our context, where the training dataset is downsampled using K-Means clustering, this added randomness helps compensate for the reduced data diversity by encouraging the model to generalize beyond the discrete cluster centroids.

Finally, the sampled value \( \hat{x}_t \) is then forwarded to the critic, which estimates the Wasserstein distance between real and generated samples. The generator and critic are trained adversarially to improve the fidelity of synthetic predictions under normal operating conditions. This structure allows the generator to perform probabilistic, context-aware forecasting of the next time step, forming the core of our anomaly detection mechanism. Figure~\ref{fig:model-overview} provides an overview the QWGAN work flow.

\subsection{Training Objectives}
\label{sec:training-objectives}
\paragraph{Training Objective}  
To train our quantum-enhanced generative model, we design a composite loss function comprising three complementary terms:

\begin{itemize}
  \item \textbf{Critic Score:}  
  A Wasserstein loss encourages the generator to produce samples \(\hat{x}_{t+1}\) that closely resemble real data \(x_{t+1}\) by minimizing the critic's ability to distinguish between them:
  \[
  \mathcal{L}_{\mathrm{critic}} = \mathbb{E}[D(\hat{x})] - \mathbb{E}[D(x)].
  \]
  This adversarial objective improves the fidelity of generated predictions.

  \item \textbf{KL Divergence:}  
  \[
\mathcal{L}_{\text{KL}} = \frac{1}{2} \mathbb{E} \left[ \exp(\log \sigma^2) + \mu^2 - 1 - \log \sigma^2 \right]
  \]

This regularization serves as a statistical reference baseline and encourages the learned distribution to remain close to a known, unimodal prior. In our anomaly detection context, it helps shape a stable representation of normal behavior — deviations from which can later signal anomalies.

Empirically, we observed that including the KL term significantly improved training stability, especially during the early stages of optimization. This effect is particularly important in our setting, where training is performed on quantum circuit simulators with high computational cost. In such environments, where each epoch is expensive, stabilizing the optimization trajectory is critical for efficient convergence. 

  \item \textbf{Variance Penalty:}  
  While the KL term promotes a variance near 1, this can lead to overly diffuse predictions for data normalized in \([0, 1]\). We thus include a regularization term:
  \[
  \mathcal{L}_{\mathrm{var}} = \mathbb{E}\left[\exp(\log\sigma^2)\right],
  \]
  which encourages the generator to remain confident (i.e., to keep uncertainty low) in non-anomalous contexts. The dynamic between the KL divergence and the variance penalty is tuned by the hyperparameter \(\lambda_{\mathrm{kl}}\), which allows us to control the modeled variance of normal behavior and modulate the trade-off between flexibility and certainty.
\end{itemize}

The full generator loss is given by:
\[
\mathcal{L}_{\mathrm{gen}} = -\mathbb{E}[D(\hat{x})] + \mathcal{L}_{\mathrm{var}} + \lambda_{\mathrm{kl}} \mathcal{L}_{\mathrm{KL}},
\]
 This composite loss enables the model to generate realistic, low-uncertainty predictions during normal operation, while establishing a structured probabilistic baseline suitable for detecting anomalies.

\paragraph{Critic Architecture and Objective}  
The critic network is designed to estimate the Wasserstein distance between real and generated samples, guiding the generator to produce more realistic outputs. Both the real future observation \(x_{t+1}\) and the generated prediction \(\hat{x}_{t+1}\) are passed through identical parallel pipelines composed of a QGRU followed by a HQL. These pipelines produce latent representations which are evaluated to yield scalar scores. The critic is trained to maximize the difference between these scores, following the Wasserstein GAN framework:
\[
\mathcal{L}_{\mathrm{critic}} = \mathbb{E}\bigl[D(\hat{x})\bigr] - \mathbb{E}\bigl[D(x)\bigr].
\]
To enforce the Lipschitz continuity required by the Wasserstein objective, we incorporate a gradient penalty term as proposed in WGAN-GP~\cite{gulrajani2017improved}. Specifically, we penalize the squared deviation of the gradient norm from 1 over random interpolations \(\tilde{x}\) between real and fake data:
\[
\mathcal{L}_{\mathrm{gp}} = \lambda_{\mathrm{gp}} \,\mathbb{E}_{\tilde{x}} \left[ \left( \|\nabla_{\tilde{x}} D(\tilde{x})\|_2 - 1 \right)^2 \right].
\]
The total critic loss thus becomes:
\[
\mathcal{L}_{\mathrm{total\_critic}} = \mathcal{L}_{\mathrm{critic}} + \mathcal{L}_{\mathrm{gp}}.
\]
This gradient penalty stabilizes training by ensuring the critic satisfies the Lipschitz condition.
\subsection{Anomaly Detection}

To identify anomalies in time-series sequences, our approach employs a two-stage detection strategy. At each time step, the model outputs a predicted mean and variance, which are then used to derive multiple diagnostic signals: (i) a reconstruction-based top-$k$ error, (ii) a Wasserstein critic score, and (iii) an interval-based uncertainty score.

In the first stage (Step 1), we apply an interval-based gating mechanism to determine whether further anomaly evaluation is warranted. Only when the observed value significantly deviates from the model’s confidence interval do we proceed to compute additional metrics.

The second stage (Step 2) involves calculating and combining the remaining diagnostic components to produce a final anomaly score. This gated scoring process helps reduce false positives and avoids unnecessary computation in regions where the model remains confident.

The following subsections detail each part of this process: (a) the interval violation gate, (b) the multi-metric scoring signals used for anomaly assessment, and (c) the final anomaly score derived from their combination.

\paragraph{Step 1 – Interval Violation Gating}

Let the model’s predicted mean and standard deviation at time step $t$ be $\boldsymbol{\mu}_t \in \mathbb{R}^d$ and $\boldsymbol{\sigma}_t \in \mathbb{R}^d$, and let the ground-truth observation be $\mathbf{x}_t \in \mathbb{R}^d$. For each feature $j \in \{1, \dots, d\}$, we define the per-feature violation score as:

\[
v_t^{(j)} = \max\left(0,\; |x_t^{(j)} - \mu_t^{(j)}| - \kappa \cdot \sigma_t^{(j)} \right)
\]

where $\kappa$ is a tunable hyperparameter controlling the size of the confidence interval. For instance, setting $\kappa = 2$ corresponds to a 95\% confidence interval under the Gaussian assumption.

To obtain a robust interval-based uncertainty score, we compute the average of the top-$k$ largest per-feature violations at time $t$:

\begin{equation}
S_{\text{iv}}(t) = \frac{1}{k} \sum_{i=1}^{k} \text{TopK}_i\left( \left\{ v_t^{(j)} \right\}_{j=1}^{d} \right)
\label{eq:interval}
\end{equation}

This formulation prioritizes the most significant deviations while suppressing minor noise.

The interval violation score quantifies how far the observed input deviates from the model’s predicted confidence bounds. Intuitively, it reflects how "surprised" the model is when confronted with the actual input, given what it expected under normal behavior. A low score indicates that the input lies within plausible variation, while a high score signals statistical improbability. Thus, this score serves as a lightweight, uncertainty-aware gating mechanism that filters out benign fluctuations before invoking more computationally intensive diagnostics.

\paragraph{Step 2 – Multi-Metric Investigation}


When the interval gate is activated, we compute two complementary signals:

\begin{itemize}

\item \textbf{Wasserstein Critic Score:}
\begin{equation}
\mathcal{S}_{\mathrm{critic}}(t) = -D(\hat{x}_t)
\label{eq:critic}
\end{equation}
 where a lower critic score implies a sample appears more fake and thus more anomalous.

  \item \textbf{Reconstruction Top-$k$ Error Score:}

We compute the squared error between the predicted mean $\boldsymbol{\mu}_t$ and the true value $\mathbf{x}_t$ at time $t$ across all features. To focus on the most significant deviations, we retain only the top-$k$ largest squared errors and average them:

\begin{equation}
S_{\text{top-}k}(t) = \frac{1}{k} \sum_{i=1}^{k} \text{TopK}_i\left( \left\{ \left(x_t^{(j)} - \mu_t^{(j)}\right)^2 \right\}_{j=1}^{d} \right)
\label{eq:reconstruction}
\end{equation}

This score emphasizes dominant deviations while reducing the influence of small, potentially noisy fluctuations.

\end{itemize}

\paragraph{Final Anomaly Score}

The final anomaly score is computed by summing the reconstruction-based and critic-based components:

\[
A(t) = S_{\text{top-}k}(t) + S_{\text{critic}}(t)
\label{eq:final_score}
\]

Here, $S_{\text{top-}k}(t) \in [0, \infty)$ is always non-negative, while $S_{\text{critic}}(t)$ may be positive or negative depending on the critic's output. 

To ensure comparability between the reconstruction error and the critic score, we normalize both $S_{\text{top-}k}(t)$ and $S_{\text{critic}}(t)$ to the range $[0, 1]$ using min-max normalization computed over the validation set. The final anomaly score is then given by:

\begin{equation}
A(t) = \tilde{S}_{\text{top-}k}(t) + \tilde{S}_{\text{critic}}(t)
\label{eq:final_score}
\end{equation}

where $\tilde{S}_{\text{top-}k}$ and $\tilde{S}_{\text{critic}}$ denote the normalized forms of the respective metrics. This normalization ensures that both signals contribute proportionally to the final score, preventing any single component from dominating due to scale differences.

In practice, we observed that combining these two signals enhances the detection of anomalous patterns by capturing both structural deviation and adversarial detectability.

This scoring is invoked only when the interval violation score $S_{\text{iv}}(t)$ exceeds a threshold. This conditional mechanism avoids unnecessary computation during normal behavior and focuses diagnostics only when the model expresses high uncertainty about its prediction.

\section{Results}\label{sec:results}

\subsection{Experimental Setup}

\paragraph{Quantum Simulation Configuration.}
All quantum components in our model, including the QGRU gates and Hybrid Quantum Layers (HQLs), are simulated using \textbf{PennyLane} with the \texttt{lightning.qubit} simulator for accelerated training and efficient quantum circuit evaluation. Each Variational Quantum Circuit (VQC) operates on \textbf{6 qubits}, balancing between expressive power and manageable simulation cost. 

The proposed model was trained on both noiseless and noisy simulators, and subsequently tested on all three environments—the noiseless simulator, the noisy simulator, and real IBM Quantum hardware.

To emulate realistic quantum noise during training, single- and two-qubit error channels were randomly introduced into the Variational Quantum Circuit (VQC). The noise model was implemented through stochastically applying Pauli errors (\texttt{X}, \texttt{Y}, \texttt{Z}) or flipped control–target CNOT operations with probabilities $p=0.1$ and $p=0.2$, respectively. These perturbations are inserted after data encoding and parameterized rotations within each layer of the circuit to simulate decoherence and gate infidelities typical of noisy intermediate-scale quantum (NISQ) devices. By incorporating such stochastic noise during optimization, the model learns parameters that are inherently robust to quantum noise, facilitating improved generalization when transferred to real quantum hardware.

\paragraph{Time Window Size Estimation.}

To determine an appropriate sliding window size $w$, we analyze statistics from the validation set. Since the attack periods in the validation data are generally short and infrequent, naively averaging durations would lead to overly small window sizes that fail to capture stable temporal patterns. To address this, we compute a \textbf{probability-weighted geometric mean} that balances short attack intervals against longer normal periods:

\[
w = \mu_d^p \cdot g^{1-p}
\]

Here, $\mu_d$ represents the average duration of labeled attack intervals, computed from ground-truth annotations. The value of $g$ corresponds to the average time between successive attacks (i.e., normal intervals). The probability $p$ reflects the proportion of time the system is under attack, estimated by dividing the total number of attack-labeled time steps by the overall sequence length. These statistics are derived empirically from the validation set.

\paragraph{Thresholding Mechanism.}
To improve robustness and temporal adaptability, we apply a local, adaptive thresholding strategy at each time step $t$. A sliding window $W_t$ is centered around the current index, and we compute the mean and standard deviation of anomaly-related scores within that window. The adaptive threshold $T_t$ is then defined as:

\[
T_t = \mu(W_t) + k \cdot \sigma(W_t)
\]

where $k$ is a tunable sensitivity hyperparameter. This threshold is used in two places within our detection pipeline:

\begin{itemize}
    \item First, it serves as a gate: we invoke deeper diagnostics (i.e., compute $A(t)$) only if the interval violation score $S_{\text{iv}}(t)$ exceeds $T_t$.
    \item Second, it is used for final decision-making: an anomaly is flagged if the computed final score also satisfies $A(t) > T_t$.
\end{itemize}

This dual usage enables both efficient screening and adaptive alerting, ensuring that anomalies are flagged relative to the local behavior of the system.
\paragraph{Hyperparameter Selection}

To ensure reproducibility and interpretability, we report in Table~\ref{tab:hyperparams-final} the main hyperparameters used in our model, along with their description.

\begin{table*}[h]
\caption{Hyperparameters Used in Our Model}
\label{tab:hyperparams-final}
\centering
\begin{tabular}{p{3.8cm}p{2cm}p{10.0cm}}
\toprule
\textbf{Hyperparameter} & \textbf{Value} & \textbf{Description} \\
\midrule
Learning rate ($\alpha$) & 0.001 & Used for training both generator and discriminator via gradient descent \\
Window size ($\tau$) & 3 & Number of past timesteps used as conditional input for QGAN \\
Number of clusters ($n$) & 300 & KMeans clusters used for successive data injection (SuDaI) \\
Rotation encoding & $\arccos(x)$ & Maps scaled input $x$ to quantum gate rotations \\
Ansatz depth & 12 & Total number of alternating (encoding + variational) layers used in the Variational Quantum Circuit (VQC), controlling overall model expressiveness \\
Ansatz layers (SuDaI) & 6 & Number of ansatz blocks where new timestamps are injected using the SuDaI technique; among the 12 total layers, data is injected into 6 of them \\

Number of qubits & 5 (G), 5 (D) & Generator uses 5 qubits; discriminator uses 5 qubits \\
Input scaling range & $[0, 1]$ & Range to which raw features are scaled before encoding \\
$k$ (Thresholding mechanism) & 1.5 & Number of windows used to compute adaptive threshold score \\

$p$ (attack probability) & $\frac{A \cdot \mu_d}{N}$ & Estimated proportion of time the system is under attack; $N$ is total number of timesteps; $A$ is the total number of attacks; $\mu_d$ Average duration of labeled attack intervals \\
$g$ (normal interval) & $\frac{N}{A} - \mu_d$ & Average time between successive attacks (i.e., average benign interval) \\

Top-$k$ (Interval Violation Gating) & 3 & Number of top features used in Interval Violation Gating \\
Top-$k$ (Reconstruction error) & 3 & Number of most significant dimensions used to compute final score \\
Feature selection count & 16 & Number of selected features from Gini method \\
\bottomrule
\end{tabular}
\end{table*}

\subsection{Evaluation Metrics}

Evaluating time-series anomaly detection models requires metrics that accurately reflect both detection correctness and the temporal consistency of detected anomaly intervals. Traditional point-wise metrics such as precision, recall, and F1-score fail to capture the sequential structure of anomalies, which are typically characterized by sustained abnormal behavior over continuous intervals rather than isolated points.

To address this limitation, we adopt the enhanced time-series aware evaluation framework introduced in~\cite{hwang2022overrate}, which extends conventional precision and recall to better account for partial and fragmented detections in time-series contexts. Specifically, the \textit{enhanced time-aware precision} (\textbf{eTaP}) rewards consistent detection of anomalous intervals, while the \textit{enhanced time-aware recall} (\textbf{eTaR}) penalizes fragmented or incomplete coverage of anomaly ranges. 
We combine these two components into a harmonic mean to compute the overall time-series aware F1 score, which we refer to as \textbf{TaF1}:

\[
\text{TaF1} = \frac{2 \cdot \text{eTaP} \cdot \text{eTaR}}{\text{eTaP} + \text{eTaR}}.
\]

This combined metric offers a more realistic assessment of anomaly detection performance in scenarios where anomalies unfold over extended periods, such as in industrial control systems.

\subsection{Results}

Table~\ref{tab:hai-taf1} presents the performance of our proposed model
compared to several baselines on the HAI dataset, evaluated using time-aware metrics.
The baseline results (RNNv1, RNNv2) are directly reported from prior work~\cite{hwang2022overrate}, and were
not retrained in our setting.

Our method achieves the highest TaF1 score of 0.89, outperforming both RNN-based baselines (0.83 for RNNv1 and
0.78 for RNNv2) as well as the naive last-timestep predictor (0.04). The latter serves as a non-learned baseline
that simply reuses the previous value as the prediction, illustrating the difficulty of the forecasting task.

To further assess the contribution of our proposed model, we introduce an additional baseline: the \textit{Cluster-Likelihood Baseline}.
This method relies solely on statistical modeling of the target distribution within each cluster obtained from KMeans
applied to training windows. At test time, anomaly detection is based on the log-likelihood of the predicted point
under the corresponding cluster’s Gaussian distribution. Despite its principled design, this baseline achieves a TaF1
of only 0.23, with a recall (eTaR) of 0.15—highlighting its limited ability to capture temporally extended anomalies.

In addition to its superior TaF1 score, our model attains an eTaP of 0.93, reflecting its prompt detection capability,
and an eTaR of 0.85, indicating strong coverage of complete anomaly intervals. These results collectively demonstrate that
our quantum-enhanced GAN significantly outperform both deep
learning and statistical alternatives.

\begin{table}[h]
\caption{Time-Aware F1 Metrics on the HAI Dataset (from~\cite{hwang2022overrate})}
\label{tab:hai-taf1}
\centering
\begin{tabular}{lccc}
\toprule
Model & TaF1 & eTaP & eTaR \\
\midrule
RNNv1 & 0.83 & 0.90 & 0.84 \\
RNNv2 & 0.78 & 0.75 & 0.69 \\
Last Timestep & 0.04 & 0.03 & 0.09 \\
Cluster-Likelihood Baseline & 0.23 & 0.48 & 0.15 \\
QWGAN (Noisy sim)  & 0.78 & 0.85 & 0.73 \\
\textbf{QWGAN (Noiseless sim)} & \textbf{0.89} & \textbf{0.93} & \textbf{0.85} \\
\bottomrule
\end{tabular}
\end{table}

\begin{figure*}[h!]
  \centering
  \includegraphics[width=\linewidth]{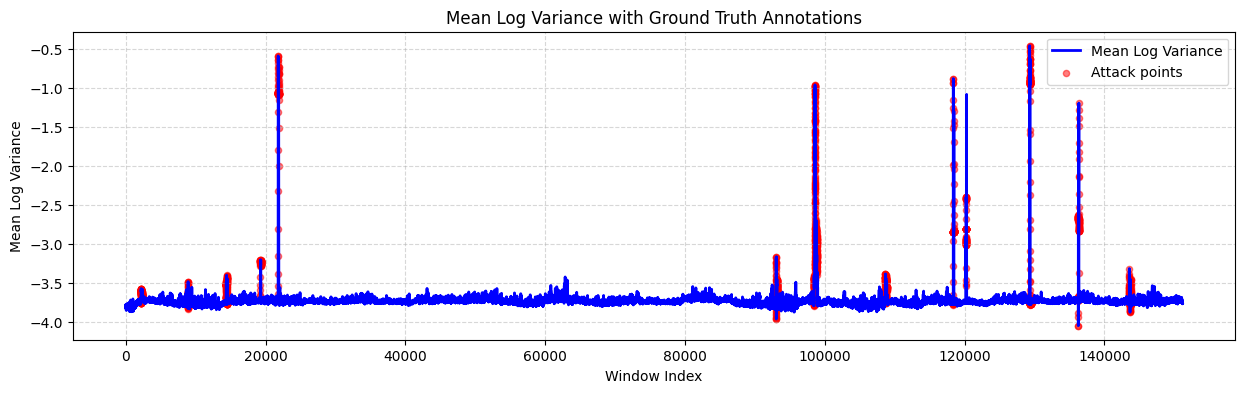}
   \caption{Scatter plot of scaled mean log‐variance, with ground‐truth anomalies highlighted by red square outlines. }
  \label{fig:logvar_scatter}
\end{figure*}

To further evaluate
the effect of quantum noise during training, a noisy-simulator
variant (QWGAN (Noisy sim)) was trained under gate error conditions. As summarized in Table~\ref{tab:hai-taf1},
this configuration achieved a TaF1 of 0.78 which is slightly lower than
the noiseless case but still competitive with classical baselines.
The modest degradation highlights the expected impact of
quantum noise while confirming the model’s inherent robustness.
Moreover, as later validated when we test the model on real quantum hardware,
the parameters optimized on the noisy simulator are expected to exhibit strong sim-to-real consistency, suggesting that the proposed architecture remains practically viable under realistic NISQ conditions for small-scale models, while acknowledging that full training on noisy quantum hardware remains an open challenge..

\subsection{Ablation Study}

The ablation results in Table~\ref{tab:ablation} highlight the contribution of each component of our composite anomaly scoring strategy (as defined in Equations~\ref{eq:interval},~\ref{eq:reconstruction}, and~\ref{eq:critic}). Specifically, we assess the impact of interval violation gating, reconstruction-based error, and Wasserstein critic feedback.

When all three components are combined into the final score (Equation~\ref{eq:final_score}), the model achieves the highest TaF1 (0.89) and the best early-time precision (eTaP = 0.93), confirming that the fusion of diverse signals leads to more timely and precise anomaly detection. Additionally, we observe an eTaR of 0.85, indicating strong recall across complete anomaly intervals.

These findings support the effectiveness of the full scoring strategy and justify the use of multi-signal integration in our final anomaly detection framework.

Removing the interval-violation gate (“Critic Only” and “Reconstruction Only” rows) lowers recall markedly (eTaR drops to 0.63 and 0.72, respectively), confirming that the gate is essential for sustaining detection across the full duration of an attack. Conversely, using \emph{only} the interval-violation score achieves an eTaR of 0.85—on par with the full model—yet loses precision (eTaP falls to 0.91) and slightly trails in TaF1. This suggests that variance-driven gating is a strong standalone detector for sustained anomalies but benefits from the critic’s realism check to reduce false positives.

Finally, the “Critic Only” configuration achieves high precision (0.85) but the lowest recall, illustrating that critic feedback excels at flagging obvious outliers yet misses subtle deviations. Taken together, these findings show that interval violation provides broad coverage, the critic sharpens precision, and their combination —augmented by reconstruction— offers the most balanced and discriminative anomaly signal.

\begin{table}[h!]
\centering
\caption{Ablation Study Results on the HAI Dataset}
\label{tab:ablation}
\begin{tabular}{lccc}
\toprule
\textbf{Configuration} & \textbf{eTaP} & \textbf{eTaR} & \textbf{TaF1} \\
\midrule
Full Model      & \textbf{0.93} & \textbf{0.85} & \textbf{0.89} \\
Critic Only         &   0.85 & 0.63 & 0.77 \\
Reconstruction error Only & 0.85 & 0.72 & 0.78 \\
Interval Violation Only & 0.91 & \textbf{0.85} & 0.88 \\
\bottomrule
\end{tabular}
\end{table}

To better understand the role of predictive variance in our model, we visualize in Figure~\ref{fig:logvar_scatter} the evolution of $\log(\sigma^2)$ across time. The plot reveals a clear correlation between elevated log-variance values and anomalous events. Specifically, during anomalous intervals, the generator consistently outputs higher uncertainty, reflected by spikes in $\log(\sigma^2)$. This suggests that the model becomes less confident in its predictions when exposed to inputs that diverge from the learned distribution of normal behavior.

This behavior validates the intuition behind penalizing the variance in our generator loss: by discouraging excessive uncertainty on normal data, we encourage sharper, more confident predictions in the expected regime. As a result, sudden increases in variance can be interpreted as early indicators of distributional shift, thus improving anomaly localization. These findings support the use of variance as an informative signal for anomaly detection at inference time.

\subsection{Real Quantum Hardware Results}
To further evaluate the deployability of the proposed QGRU-WGAN on current noisy intermediate-scale quantum (NISQ) devices, the trained generator–critic pair was executed on real IBM Quantum hardware. The parameters optimized on the simulator were directly transferred without retraining.  
Due to hardware execution constraints, only a representative fragment of the test sequence containing a known attack was used for on-device validation. This ensures that the observed results are comparable to simulator-based testing while maintaining feasibility under limited shot and queue resources.

\begin{figure*}[t]
    \captionsetup[subfloat]{font=small, labelfont=bf}
    \centering
    \subfloat[Model trained and tested on noiseless simulator (reference baseline).]{%
        \includegraphics[width=0.48\textwidth]{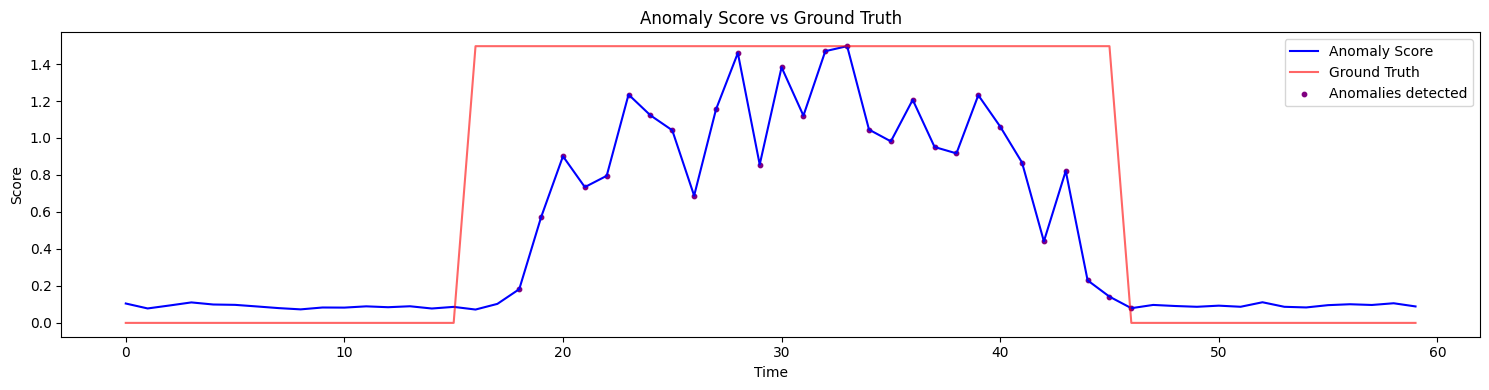}%
        \label{fig:sim_noiseless}} \hfill
    \subfloat[Model trained on noiseless simulator; tested on real hardware.]{%
        \includegraphics[width=0.48\textwidth]{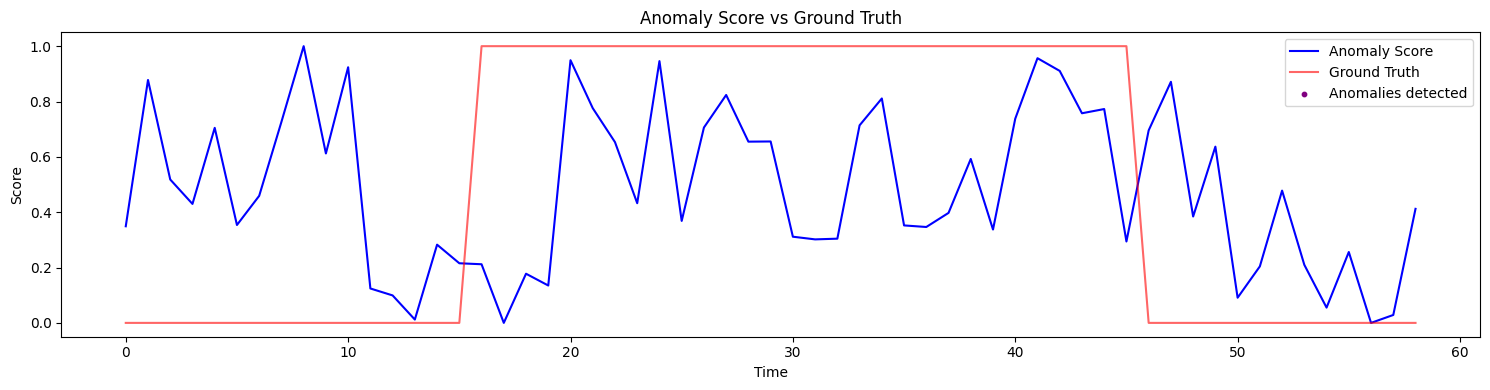}%
        \label{fig:hw_noiseless_train}}

    \subfloat[Model trained on noisy simulator; tested on real hardware.]{%
        \includegraphics[width=0.48\textwidth]{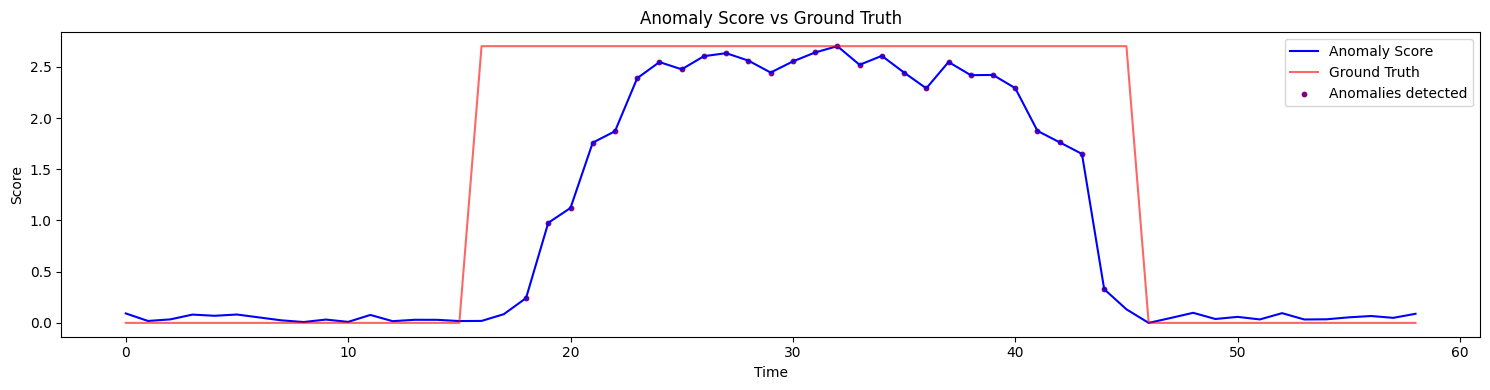}%
        \label{fig:hw_noisy_train}} \hfill

    \caption{Anomaly score visualizations under different training–testing configurations. Each plot shows the detected anomaly scores (blue) against ground-truth events (red). The model trained on a noisy simulator exhibits strong sim-to-real transfer.}
    \label{fig:real_quantum_results}
\end{figure*}

\begin{table}[h]
\centering
\caption{Sim-to-real anomaly detection performance comparison}
\label{real-harware-results-table}
\begin{tabular}{lccc}
\toprule
\textbf{Metric} & \textbf{eTaP} & \textbf{eTaR} & \textbf{TaF1} \\
\midrule
Real hardware (trained on noisy sim) & 1.000 & 0.950 & 0.974  \\
Noiseless simulator (in-domain eval.) & 1.000 & 0.983 & 0.992  \\
\bottomrule
\end{tabular}
\end{table}

Figure~\ref{fig:real_quantum_results} presents the anomaly scores under the three main evaluation settings. 
The reference configuration (Fig.~\ref{fig:sim_noiseless})—a model trained and tested on the noiseless simulator—represents the ideal upper-bound scenario. This case establishes the expected anomaly score profile under error-free execution.

When the same noiseless-trained model is executed on real hardware (Fig.~\ref{fig:hw_noiseless_train}), the detection capability collapses: the anomaly curve shows no correlation with ground-truth events, and the signal is dominated by device-induced fluctuations. This mismatch confirms that training on idealized simulators fails to generalize to real NISQ environments.

In contrast, the model trained on a noisy simulator (Fig.~\ref{fig:hw_noisy_train}) successfully reproduces anomaly patterns on hardware, showing strong agreement with the baseline simulator performance (Table~\ref{real-harware-results-table}). The inclusion of simulated noise during training allows the quantum parameters to encode stable, transferable features resilient to physical noise.  
This experiment demonstrates that noise training effectively bridges the simulation-to-hardware performance gap, achieving a near-identical time-series aware F1 (0.974 vs.\ 0.992) while maintaining perfect precision.  
Overall, the results validate that the proposed QGRU-WGAN can operate reliably on real quantum processors when trained under realistic noise conditions.

\section{Conclusion}\label{sec:conclusion}

In this paper, we introduced a novel quantum-enhanced anomaly detection framework that integrates a Quantum Gated Recurrent Unit (QGRU) with a hybrid adversarial training scheme. Our architecture leverages the expressiveness of Variational Quantum Circuits (VQCs) and SuDaI for sequence modeling trick to generate uncertainty-aware predictions. To improve training stability and model reliability, we incorporated a KL-divergence regularizer and a log-variance penalty alongside a Wasserstein-based critic with gradient penalty. Furthermore, we proposed an interpretable multi-stage anomaly scoring mechanism that begins with interval violation analysis leading to better scoring based on critic evaluation and reconstruction error. Experimental  results demonstrated the model's strong performance across extended time-aware metrics (TaF1: 0.89, eTaR: 0.85), validating the utility of our quantum-inspired design and scoring strategy.

In addition, the model was successfully deployed on real IBM Quantum hardware without retraining. The hardware evaluation confirmed that parameters learned under simulated noise transfer effectively to physical qubits, demonstrating strong sim-to-real consistency. This validation highlights the practical feasibility of deployement of hybrid quantum neural architectures for real-world anomaly detection tasks.

This work marks an interesting step toward scalable and explainable quantum machine learning systems for network anomaly detection, combining quantum expressiveness with principled uncertainty modeling.

\printbibliography
\end{document}